\documentclass[11pt,twocolumn]{article}

\usepackage{geometry}
\usepackage{subfig}
\usepackage{indentfirst}    
\usepackage{tabularx} 
\usepackage{graphicx}       
\graphicspath{{./}{fig/}}   
\usepackage{float}          
\usepackage{caption}        
\usepackage{booktabs}       
\usepackage[section]{placeins} 
\usepackage{balance}   
\usepackage[UTF8,fontset=fandol]{ctex}     
\usepackage{enumitem}       
\usepackage{amsmath}        
\usepackage{multirow}       
\usepackage{cuted}          

\usepackage{xcolor}
\usepackage{tcolorbox}
\tcbuselibrary{listings, breakable, skins}
\usepackage{listings}

\definecolor{promptTitle}{RGB}{81,125,150}
\newtcolorbox{promptbox}[1]{%
    title={#1},
    colback=white,
    colframe=promptTitle,
    colbacktitle=promptTitle,
    coltitle=white,
    fonttitle=\bfseries,
    enhanced,
    boxrule=1.0pt,
    arc=1.2mm,
    left=6pt,right=6pt,top=6pt,bottom=6pt
}

\lstdefinestyle{promptlisting}{%
    basicstyle=\normalfont\small,
    breaklines=true,
    columns=fullflexible,
    keepspaces=true,
    showstringspaces=false
}

\makeatletter
\@ifundefined{refname}{}{}
\@ifundefined{bibname}{}{}
\makeatother

\usepackage{dblfloatfix}
\usepackage[colorlinks=true, linkcolor=black, urlcolor=blue, citecolor=black, pdfborder={0 0 0}]{hyperref}
\usepackage{cleveref}       

\title{\textbf{Legal-DC: Benchmarking Retrieval-Augmented Generation for Legal Documents}}

\author{
Yaocong Li$^{a,\dagger}$ \qquad Qiang Lan$^{a,\dagger}$ \qquad Leihan Zhang$^{a,*}$ \qquad Le Zhang$^{b}$
}

\date{}  

\newcommand{\keywords}[1]{\noindent\textbf{Keywords:} #1\par\vspace{1em}}

\begin{document}

\twocolumn[
\begin{@twocolumnfalse}
\maketitle

\begin{center}
\footnotesize
$^{a}$School of Economics and Management, Beijing University of Posts and Telecommunications, No. 10 Xitucheng Road, Haidian District, Beijing, 100876, China\\
$^{b}$College of Computing, Beijing Information Science and Technology University, No. 12 Xiaoying East Road, Haidian District, Beijing, 100085, China\\
$^{\dagger}$These authors contributed equally to this work\\
$^{*}$Corresponding author: \texttt{zhangleihan@gmail.com}
\end{center}

\vspace{0.5em}
\hrule height 0.8pt
\vspace{0.5em}

\begin{abstract}
\noindent Retrieval-Augmented Generation (RAG) has emerged as a promising technology for legal document consultation, yet its application in Chinese legal scenarios faces two key limitations: existing benchmarks lack specialized support for joint retriever-generator evaluation, and mainstream RAG systems often fail to accommodate the structured nature of legal provisions. To address these gaps, this study advances two core contributions: First, we constructed the Legal-DC benchmark dataset, comprising 480 legal documents (covering areas such as market regulation and contract management) and 2,475 refined question-answer pairs—each annotated with clause-level references—filling the gap for specialized evaluation resources in Chinese legal RAG. Second, we propose the LegRAG framework, which integrates legal adaptive indexing (clause-boundary segmentation) with a dual-path self-reflection mechanism to ensure clause integrity while enhancing answer accuracy. Third, we introduce automated evaluation methods for large language models to meet the high-reliability demands of legal retrieval scenarios. LegRAG outperforms existing state-of-the-art methods by 1.3\% to 5.6\% across key evaluation metrics. This research provides a specialized benchmark, practical framework, and empirical insights to advance the development of Chinese legal RAG systems. Our code and data are available at \url{https://github.com/legal-dc/Legal-DC}.
\end{abstract}

\vspace{0.5em}
\vspace{0.5em}
\hrule height 0.8pt
\vspace{1em}
\end{@twocolumnfalse}
]

\section{Introduction}

Large Language Models (LLMs) have demonstrated exceptional capabilities in dealing with professional document Q\&A problems in medicine, finance, and law. Individuals can use domain-specific corpora to fine-tune the models  and enhance their vertical domain expertise. However, LLMs still face issues like hallucinations \cite{6} and outdated corpora.

The Retrieval-Augmented Generation (RAG) approach \cite{2}, which retrieves and integrates relevant information from reliable external sources, effectively mitigates the issues. Consequently, the application of RAG has rapidly evolved in areas like medicine \cite{3}, finance \cite{4}, and law \cite{5}. 

A complete RAG consists of a corpus, a retriever, and a generator, with more sophisticated retrievers including components such as embedding models and reranking models \cite{1}. In the RAG process, retrieved documents are crucial for accurate answers, especially in professional fields. Poor quality documents hinder LLMs from providing satisfactory responses. The generator and retriever together determine the quality of the final answer.

In legal document consultation, large language models (LLMs) must strictly adhere to statutory provisions when answering questions, which highlights the fundamental role of accurately retrieving relevant legal documents and contents in retrieval-augmented generation (RAG) systems. However, existing legal RAG benchmarks mainly focus on answer generation and lack dedicated evaluation of retrievers, largely due to the absence of fine-grained annotations that explicitly link questions to their corresponding reference contents within legal documents. To address this limitation, we construct a benchmark dataset for legal document consultation in the context of market business activities in China, named Legal-DC, which enables joint evaluation of both retrievers and generators in RAG systems. Legal-DC consists of 2,475 question–answer–document entries derived from 480 legal documents, including laws, administrative regulations, and departmental rules governing market entities in China, with two rounds of manual annotation conducted to ensure annotation reliability. Furthermore, we propose a simple yet effective benchmark framework based on a classic RAG architecture that adopts a parallel document processing strategy integrating chunk-based semantic segmentation and article-level structural segmentation, allowing legal documents to be represented and retrieved at multiple granularities simultaneously. Experimental results on Legal-DC demonstrate that this parallel chunk–article processing strategy, together with reranking models, significantly improves the performance of legal RAG systems, while also revealing that LLMs are sensitive to noisy retrieved documents, which may lead to incorrect legal references in generated answers. In summary, our contributions are as follows:

\begin{enumerate}[label=(\arabic*),noitemsep,topsep=0pt,leftmargin=*]
  \item We construct a benchmark dataset (Legal-DC) for comprehensively evaluating the retrievers and generators of Chinese legal RAG systems, and develop a toolkit that facilitates the combination and testing of different components (including retrievers and generators) to aid in searching for the optimal combination.
  \item We propose a concise and efficient benchmark methodology for legal document consultation RAG, demonstrating superior capabilities compared to existing mainstream RAG systems and platforms. This includes approaches such as entity retrieval, reflection, and multi-way recall, whose experimental results provide crucial insights for enhancing RAG system performance.
\end{enumerate}

\section{Related Work}

\subsection{Retrieval-Augmented Generation}

RAG \cite{2} is a technology that combines information retrieval with generative artificial intelligence. It retrieves relevant information from an extensive document database before generating the content. The information is used to guide the generation process and enhance the accuracy and relevance of the answer. RAG effectively mitigates the hallucination issues that may arise in LLMs when generating content, accelerates the speed of knowledge updates, and strengthens the traceability of content generation. The paradigm of RAG can be divided into three categories: Naive RAG, Advanced RAG, and Modular RAG \cite{1}. The commonality among these types is that they comprise a retriever and a generator. Consequently, most researchers focus on improving the performance of RAG from these two aspects, such as fine-tuning the embedding model in the retriever \cite{16,17}, mixing multiple retrieval methods for multi-path recall \cite{15}, reranking \cite{14}, etc., and fine-tuning the LLM in the generator, as well as using different thought chain methods for different scenarios \cite{18,19,20}. As the RAG method continues to be optimized, various benchmarks are also being refined, such as general benchmarks CRUD-RAG \cite{7}, ARES \cite{8}, RAGAS \cite{9}, and some domain-specific RAG benchmarks, such as FinanceBench \cite{10}, LegalBench \cite{11}, MedExpQA \cite{12}, MIRAGE \cite{13}. Currently, there is no comprehensive benchmark for RAG testing in the Chinese legal Q\&A domain. Thus, this paper introduces a Chinese legal RAG benchmark called Legal-DC, which thoroughly assesses both the retriever and the generator.

\subsection{LLMs on Legal Domains}

Researchers initially used LLMs with specific prompts to evaluate contract legal issues and compared their performance to junior lawyers. Studies showed that LLMs were feasible for contract review and more efficient, providing faster and more accurate results at a lower cost\cite{22}. Subsequently, researchers began to fine-tune LLMs with instructions to focus on specific domains, such as SaulLM-7B\cite{23}, a large language model with 7 billion parameters designed to understand and generate legal text. Nevertheless, the issues of hallucinations \cite{6} and outdated corpus remain severe concerns for the rigorous legal field, which the RAG method has effectively mitigated. Lawyer LLaMA \cite{21} is a legal RAG framework developed based on the fine-tuning of legal documents with the LLaMA model, while ChatLaw \cite{5} is a RAG framework based on the Ziya-LLaMA-13B model, which has been fine-tuned on legal documents. LLMs rely on accurate legal documents for reliable answers in the legal field, but evaluating document retrieval is often overlooked. The Legal-DC benchmarking system we propose enables more effective evaluation of both the accuracy with which retrieval engines acquire text and the correctness of text-based responses generated by large language models.

\section{The Benchmark Dataset}

\subsection{Dataset Construction}
Given that various market operators require frequent legal consulting services in their business operations and RAG technologies hold substantial potential to address this demand, we collected laws, rules, and regulations governing the business activities of market entities from the China National Laws and Regulations Database. The dataset comprises 480 documents in Doc or PDF format, covering key areas such as market supervision, contract management, food safety, and unmanned aerial vehicle management—all of which are closely related to the daily business operations of market entities. We constructed the benchmark dataset through three rigorous stages (\Cref{fig:dataset_construction}), with detailed operations and quality control mechanisms as follows:

\begin{figure}[!htbp]
    \centering
    \includegraphics[width=\linewidth]{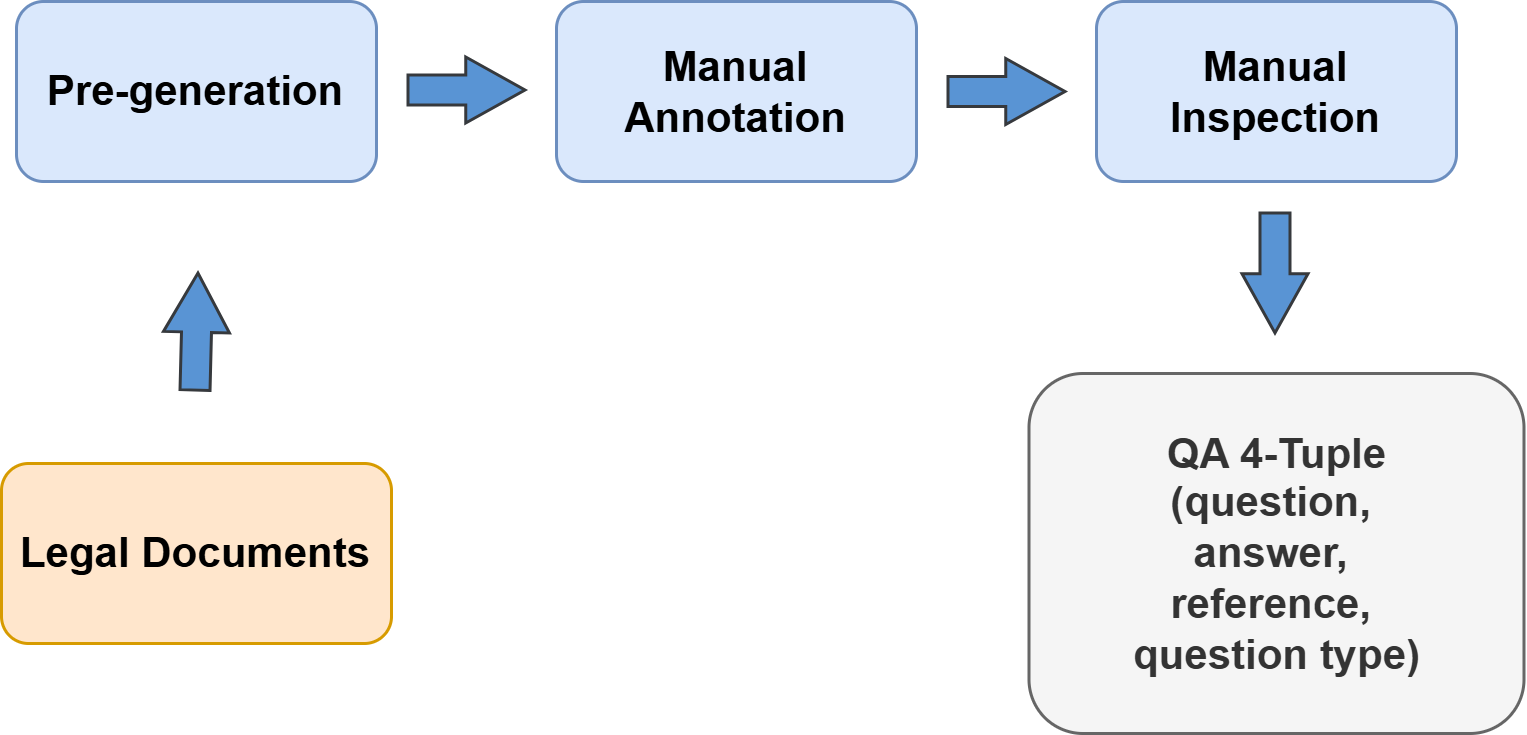} 
    \caption{Dataset construction process.}
    \label{fig:dataset_construction}
\end{figure}

\textbf{Candidate question and answer (QA) pair generation.}
First, we parsed the content of the 480 legal documents to extract structured information such as clauses, articles, and key terms, ensuring that the generated QA pairs are grounded in the document content. We then employed two advanced LLMs—GPT-3.5 and ERNIE-4.0—to initially generate candidate QA pairs for 50 representative documents. The core principle for generation is that each question must be answerable exclusively based on the content of the corresponding document, and the answers must strictly align with legal provisions without adding external knowledge.

To select the optimal generator, we conducted a manual evaluation of the generated QA pairs from both models, focusing on three metrics: (a) relevance to document content (whether the question targets key legal provisions), (b) practical significance for market operators (whether it addresses high-frequency consulting needs), and (c) answer accuracy (whether the answer is consistent with the original clauses). The results showed that ERNIE-4.0 outperformed GPT-3.5 in all three metrics (relevance: 82\% vs. 68\%, significance: 79\% vs. 63\%, accuracy: 91\% vs. 83\%). Thus, we adopted ERNIE-4.0 to generate 10 candidate QA pairs for each of the 480 documents, resulting in an initial pool of 4,800 candidate QA pairs.

\textbf{Selection and annotation.}
Due to inherent limitations of LLMs, a considerable portion of the initially generated candidate QA pairs were either of low practical significance (e.g., targeting trivial clauses irrelevant to business operations) or poor quality (e.g., ambiguous questions, answers inconsistent with legal provisions). To address this, we recruited five annotators with relevant backgrounds: three were senior undergraduate students majoring in Law (with basic knowledge of commercial law and administrative law) and two were postgraduates specializing in Legal Informatics (with experience in legal document annotation). Prior to formal annotation, all annotators participated in a 2-hour training session, which covered: (a) detailed definitions of ``significance'' and ``quality'' for QA pairs, (b) classification criteria for question types, (c) operation guidelines for reference content annotation, and (d) handling of ambiguous cases.

Each annotator was assigned 96 documents (480 documents distributed evenly among five annotators) to ensure consistent workload distribution. For each document, the annotator first read all 10 candidate QA pairs and then thoroughly reviewed the full text of the document to verify the alignment between the QA pairs and the document content. For each candidate QA pair, the annotator evaluated two core dimensions: (a) Significance—determined by whether the question reflects high-frequency legal needs of market operators (e.g., ``Is real-name registration required for purchasing micro-drones?'' is significant, while ``What is the font size of Article 5 in the document?'' is not); (b) Quality—assessed by whether the question is clear and unambiguous, and whether the answer is fully supported by the document content (no hallucinations or omissions).

To illustrate these annotation standards, we provide examples of accepted and rejected question-answer pairs. Accepted high-importance and high-quality questions: What consequences will foreign investors face for failing to report investment information on time? Answer: The commerce authority orders rectification within a specified timeframe; failure to comply results in a fine of RMB 100,000 to 500,000. Reason: Involves a high-frequency compliance issue with explicit legal consequences. Reject questions with inaccurate answers: What is the validity period of a food business license? Generated answer: 3 years; Actual regulation: 5 years. Reason: The generated answer contradicts documented regulations, exhibiting hallucination.

QA pairs that met both ``high significance'' and ``high quality'' criteria were retained. For these retained pairs, the annotator further annotated: (a) Question type—categorized into four types based on legal semantic understanding and cognitive complexity: concept explanation (targeting specialized terminology or legal concepts), summary induction (extracting key information from lengthy clauses), logical reasoning (multi-step inference based on provisions), and others (questions not falling into the above three categories); (b) Reference content—precisely locating the specific clauses or paragraphs in the document that support the answer, including article numbers (e.g., Article 6) and key sentences to ensure traceability. For documents where fewer than 4 QA pairs were retained after screening, the annotator was required to design new QA pairs based on the document's core provisions, following the same significance and quality standards, and complete the corresponding annotations. This minimum threshold of 4 QA pairs per document ensures sufficient question diversity while maintaining annotation feasibility across the 480-document corpus.

To ensure annotation quality, the project director implemented a real-time supervision mechanism: (a) randomly checking 10\% of the annotated documents per annotator (e.g., 9-10 documents per annotator) to verify the accuracy of QA pair selection, question type classification, and reference content annotation; (b) establishing a dispute resolution group to address ambiguous cases (e.g., conflicting judgments on QA pair significance) through collective discussion; (c) requiring annotators to revise problematic annotations within 24 hours of feedback. After completing the annotation task, each annotator was paid \$100 as compensation. This stage resulted in 3,120 preliminary valid QA pairs.

To provide transparency in the quality control process, we conducted a systematic analysis of the 1,680 rejected QA pairs from the initial 4,800 candidates. The rejection reasons were categorized as follows: (a) Low practical significance (38.7\%, 650 pairs)—questions targeting trivial clauses irrelevant to business operations (e.g., ``What is the effective date of this regulation?''); (b) Answer inaccuracy (29.2\%, 490 pairs)—generated answers inconsistent with legal provisions or containing hallucinations; (c) Question ambiguity (18.5\%, 311 pairs)—vague or multi-interpretable questions that cannot be definitively answered; (d) Redundancy (13.6\%, 229 pairs)—duplicated or highly similar questions within the same document. This quality filtering ensures that the final dataset focuses on legally substantive and practically valuable question-answer pairs.

\textbf{Cross proofreading and annotation.}
To mitigate subjectivity and enhance reliability, we recruited five new annotators (with the same background requirements) for cross proofreading. This stage adopted a ``blind review'' mode—annotators were not informed of the initial annotation results to avoid bias. Each document and its corresponding preliminary QA pairs were assigned to two independent second-round annotators for cross-validation. The proofreading focus included: (a) re-evaluating the significance and quality of the retained QA pairs; (b) verifying the accuracy of question type classification and reference content annotation; (c) identifying and replacing QA pairs with insufficient significance (e.g., replacing trivial questions with practical ones) and correcting annotation errors (e.g., misclassified question types, inaccurate reference content). Inconsistent judgments were resolved by a senior arbitrator (with 3 years of experience). If the number of valid QA pairs for a document was still less than four after proofreading, the second-round annotators collaboratively designed additional QA pairs to meet the minimum requirement.

Annotators were paid \$100. First-round annotation averaged 6.2 hours/person, and cross-proofreading 3.8 hours/person. The entire annotation process spanned 6 weeks, with Week 1-2 for training and first-round annotation, Week 3-4 for cross proofreading, and Week 5-6 for dispute resolution and final quality verification. Cohen's Kappa coefficients demonstrated excellent consistency: 0.83 (QA selection), 0.87 (question type), and 0.91 (reference content).

After the three stages mentioned above, the final Legal-DC dataset contains 2,475 high-quality QA entries corresponding to 480 legal documents, with each document having no fewer than four QA pairs. Each QA entry is a five-tuple structured as (document content, question, answer, question type, reference content). \Cref{tab:table1} shows token-length statistics of questions, answers, and reference content, and \Cref{tab:data_samples} presents sample entries of the dataset.

Regarding the distribution of question types: Concept explanation accounts for 10.83\% (268 entries), Summary induction accounts for 65.21\% (1,614 entries), Logical reasoning accounts for 21.33\% (528 entries), and Others account for 2.63\% (65 entries). This distribution is consistent with the actual demand for legal consultation—market operators frequently require summarization of legal provisions and logical inference based on clauses, while concept explanation needs are relatively less frequent. Additionally, the dataset covers multiple legal domains, with the top three being market supervision (32\%), contract management (28\%), and product safety (18\%), ensuring broad coverage of practical business scenarios.

\begin{table}[t]
    \centering
    \caption{Token lengths of questions, answers and reference documents in the Legal-DC benchmark.}
    \label{tab:table1}
    \begin{tabular*}{\columnwidth}{@{\extracolsep{\fill}}lccc@{}}
    \toprule
    \textbf{}  & \textbf{Min} & \textbf{Max}  & \textbf{Average} \\
    \midrule
    Query      & 6            & 79            & 21.69            \\
    Answer     & 3            & 672           & 100.05           \\
    Document   & 12           & 2006          & 132.26           \\
    \bottomrule
    \end{tabular*}
\end{table}

\subsection{Evaluation metrics}

Two groups were designed and used in the proposed benchmark: metrics for the retrieval methods and metrics for the answer generation methods.

For each question, if the target content is in the top $K$ retrieved content set, we deem that the retrieval method returns the right content. For the questions in the test dataset, Recall@K is the proportion of the number of questions with right retrieved content to the total number of questions.

MRR@K (Mean Reciprocal Rank@K) is based on Reciprocal Rank, which measures the rank of the first target content in the top $K$ retrieved content set. The mean Reciprocal Rank is the average of the Reciprocal Rank for a group of questions in the test datasets. The calculation formula is as follows:
\[
\text{MRR@}K =\frac{1}{N} \sum_{i=1}^{N} \frac{1}{\text{rank}_i} \tag{1}
\]
Where:
\begin{itemize}[noitemsep,topsep=0pt,leftmargin=*]
    \item $N$ denotes the total number of questions in the test dataset;
    \item $\text{rank}_i$ represents the rank of the first target content in the top $K$ retrieved content set for the $i$-th question;
    \item $K$ is the size of the retrieval candidate set (set to 5 in this study).
\end{itemize}

\subsubsection{Metrics for the answer generation methods}

For answer generation methods, besides the classic metrics including Accuracy, BLEU \cite{26} and ROUGE-L \cite{27}, we propose another metric named Document Selection Accuracy (DSA) to evaluate the ability of LLMs to select the correct content/document from the recalled set.

Accuracy is calculated based on the ROUGE-L metric. Given that answers in legal consultation scenarios often involve synonym rephrasing, clause restructuring, and cross-paragraph citations, a fixed threshold struggles to objectively reflect generation quality. This paper leverages high-quality, manually annotated question-answering samples from the Legal-DC dataset. It statistically analyzes the ROUGE-L F1 score distribution corresponding to correct answers on the validation set and selects the partition point that maximizes human-judgment consistency as the evaluation threshold. This strategy ensures consistency in key information coverage between generated and reference answers while reducing misjudgment risks caused by phrasing differences. Generated answers are classified as correct when their ROUGE-L score exceeds this threshold and as incorrect otherwise. Overall accuracy on the test set is calculated based on this classification.

DSA measures the capability of LLMs to select the right content from a set of retrieved documents. The calculation formula is as follows:
\[
\text{DSA} = \frac{d}{D} \tag{2}
\]
Where:
\begin{itemize}[noitemsep,topsep=0pt,leftmargin=*]
    \item $d$ is the number of questions for which the LLM selected the correct content from the retrieved set;
    \item $D$ is the number of questions that successfully recalled the correct document in the top $K$ retrieved content set ($K=5$ in this study).
\end{itemize}

\subsection{Automatic Evaluation Protocol}

To address the limitations of n-gram metrics like ``BLEU/\allowbreak ROUGE-L'' in adequately capturing semantic consistency, factual accuracy, and traceability reliability, we employ Qwen3-max as an automated evaluator and construct an end-to-end evaluation workflow based on a specific comprehensive evaluation prompt. This prompt emphasizes ``evidence-based prioritization'' and ``legal accuracy'', aligning particularly well with the high credibility demands of legal retrieval scenarios.

\subsubsection{Evaluation Pipeline}
The automatic evaluation process is constructed as a three-stage workflow to ensure credible and reproducible assessment results for the legal RAG system, which is implemented via a well-designed prompt template and large language model-based assessment:

\textbf{Input Concatenation.}
The legal question, top-$K$ retrieved reference documents, model-generated answer and gold reference answer are embedded into the predefined placeholders \{query\}, \{retrieved\_docs\}, \{generated\_answer\} and \{reference\_answer\}, forming a complete evaluation prompt for subsequent inference.

\textbf{LLM-based Automated Assessment.}
We adopt Qwen3-max as the automatic evaluator with deterministic inference configuration: $temperature=0$ and $top_p =1.0$. This setting disables stochastic sampling and ensures stable and fully reproducible evaluation outputs. The model returns structured JSON results with quantitative scores across four evaluation dimensions for each sample.

\textbf{Evaluation Metric Calculation.}
Four core scoring dimensions are defined: relevance\_score (evidential support and topical relevance), accuracy\_score (correctness of legal conclusions and facts), completeness\_score (coverage of key legal points) and fluency\_score (logical coherence and linguistic smoothness). A weighted total score ($total_score \in[0,100]$) is calculated as the sum of the four dimensional scores, and normalized to the final metric: $\text{LLM-Score} = total\_score/100$.

To ensure a structured and objective evaluation of legal answer quality, this study assigns different weights to multiple evaluation dimensions according to the professional judgment habits of legal experts and the intrinsic requirements of legal document consultation. Considering the high-risk nature of legal decision-making and the strong dependence on statutory correctness, the evaluation framework emphasizes content validity over linguistic expression. \Cref{tab:weight_rationale} presents the detailed weight allocation for each evaluation dimension along with the corresponding rationales.

\begin{table*}[t]
\centering
\caption{Weight Allocation and Rationale for Legal Answer Evaluation Dimensions}
\label{tab:weight_rationale}
\small
\setlength{\tabcolsep}{6pt}
\renewcommand{\arraystretch}{1.15}
\begin{tabularx}{\textwidth}{p{3cm} p{2cm} X}
\toprule
\textbf{Evaluation Dimension} & \textbf{Weight} & \textbf{Rationale} \\
\midrule
Accuracy & 40\% & Legal answers directly affect rights and obligations. Any deviation from statutory provisions or factual inaccuracies may cause serious legal risks, making correctness the fundamental requirement. \\
\midrule
Relevance & 30\% & Answers must align with query intent and retrieved evidence. Irrelevant content has no legal consultation value. \\
\midrule
Completeness & 20\% & Legal queries often involve multiple clauses. Completeness improves usefulness, but correctness remains more critical. \\
\midrule
Fluency & 10\% & Fluency enhances readability, but content validity outweighs linguistic quality in legal consultation. \\
\bottomrule
\end{tabularx}
\end{table*}

This weight setting ensures that LLM-Score aligns with real-world legal service priorities, avoiding overemphasis on non-critical dimensions and accurately reflecting the practical quality of legal RAG outputs.

\subsubsection{Prompt Template for Legal RAG Evaluation}

The core prompt template is designed following the principle of evidence-based legal assessment with clear grading criteria and standardized input structures, to guide the LLM to conduct objective and consistent evaluation. A concise core snippet of the evaluation prompt is presented in \Cref{fig:eval_prompt_snippet}.

\begin{figure}[t]
    \centering
    \begin{promptbox}{Core Evaluation Prompt Snippet}
\begin{lstlisting}[style=promptlisting]
You are a professional evaluation expert for legal Retrieval-Augmented Generation (RAG) systems. You need to assess the quality of the generated answer according to the given legal question, gold reference answer and retrieved legal documents.

Evaluation Criteria (Total: 100 points)
1. Relevance (30 points): Evaluate if the generated answer is supported by retrieved legal documents and relevant to the raised question.
2. Accuracy (40 points): Evaluate the correctness of legal conclusions, statutory clauses and factual statements in the answer.
3. Completeness (20 points): Evaluate if the answer fully covers all key legal points of the question.
4. Fluency (10 points): Evaluate the linguistic smoothness and logical coherence of the answer.

Input Information
Question: {query}
Gold Reference Answer: {reference_answer}
Model-generated Answer: {generated_answer}
Retrieved Legal Documents: {retrieved_docs}
\end{lstlisting}
    \end{promptbox}
    \caption{Core prompt snippet for automatic legal RAG evaluation.}
    \label{fig:eval_prompt_snippet}
\end{figure}

\begin{figure}[t]
    \centering
    \begin{promptbox}{Answer Generation Prompt}
\begin{lstlisting}[style=promptlisting]
You are a legal Q&A consulting expert. First, you need to select the documents related to the [question]
from the [retrieval results] and output [related documents], and then complete the [answer] according to the
[related documents].
[retrieval results]
-{reference}
[question]
{query}
[related documents]
Output the relevant documents from the [retrieval results] according to query
[answer]
Output the answer to query according to the [related documents]
\end{lstlisting}
    \end{promptbox}
    \caption{The prompt for answer generation.}
    \label{fig:prompt_answer_generation}
\end{figure}

\subsubsection{Experimental Setup and Credibility Validation}
All automatic evaluation experiments are conducted with unified and reproducible configurations to ensure the validity of the comparative results, and the credibility of LLM-based automatic evaluation is further verified by human review:
\begin{itemize}
    \item The retrieval candidate size is set as $k=5$, and the evaluation is conducted on a benchmark test set containing $\textbf{2,475}$ legal question-answer samples;
    \item Each sample is independently assessed by Qwen3-max, with an average inference latency of 1.2 seconds per sample;
    \item A random sample of 30\% test data is selected for manual expert review. The Spearman correlation coefficient $\rho=0.87$ is achieved between the LLM-based scores and human ratings, which demonstrates a strong positive correlation and validates the reliability of the automatic evaluation results.
\end{itemize}

\noindent\textit{Note}: If Qwen3-max is unavailable, alternative large language models with equivalent capability (e.g., GPT-4, Claude-3-Opus) can be adopted. It is recommended to recalibrate the evaluation threshold in a pre-experiment, and the core conclusions of this study remain unchanged.

\section{Experimental Design}

\begin{figure*}[t]
    \centering
    \includegraphics[width=\textwidth]{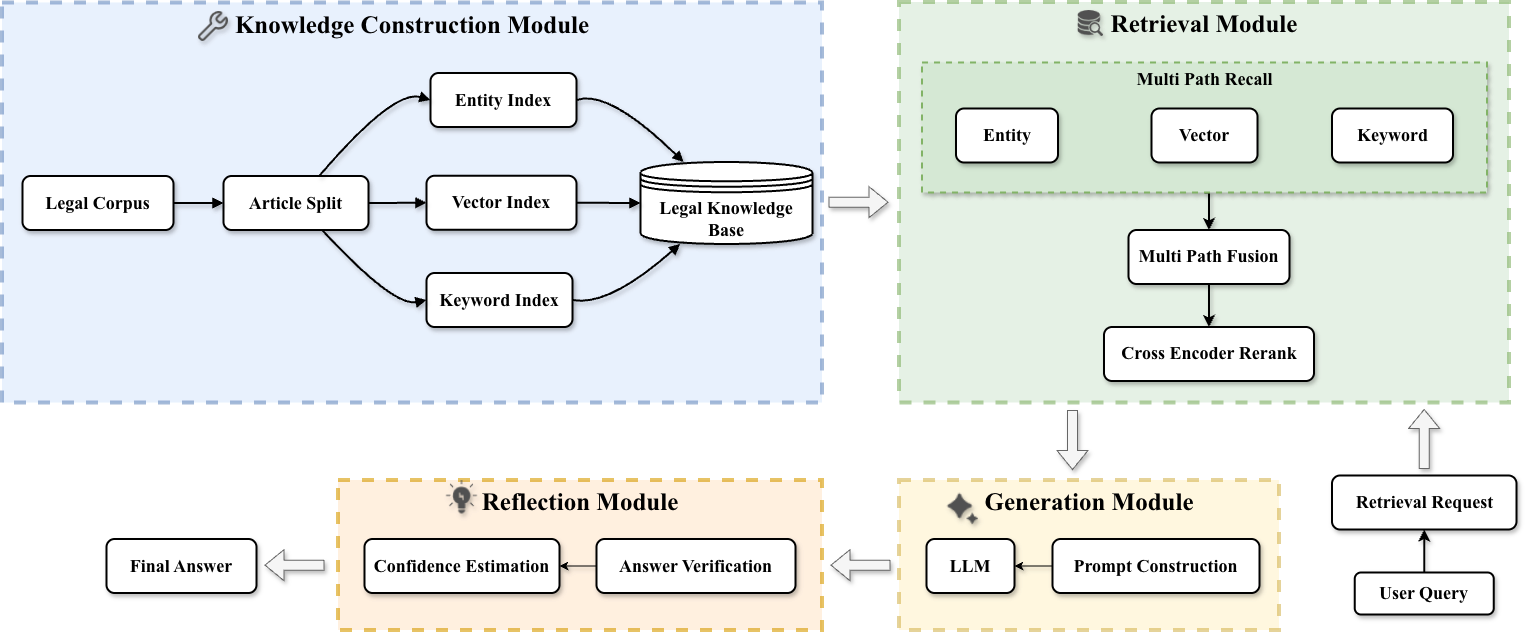}
    \caption{Architecture of the LegRAG framework.}
    \label{fig:legrag_architecture}
\end{figure*}
To formalize the Legal-DC Retrieval-Augmented Generation (RAG) framework (i.e., LegRAG, corresponding to \Cref{fig:legrag_architecture}), this section first defines core symbols and the overall objective, then elaborates on the modular implementation logic, ensuring full alignment with the functional modules and data flow in the framework diagram.

\subsection{Formal Definition of the RAG Framework}
The core components and mathematical definitions of the Legal-DC RAG framework (LegRAG) are as follows, with all symbols matching the module functions in \Cref{fig:legrag_architecture}:
\begin{itemize}
    \item $q \in Q$: A set of legal consultation queries, where $Q = \{q_1, q_2, \dots, q_{2475}\}$ (corresponding to 2475 QA pairs in the dataset).
    \item $D = \{d_1, d_2, \dots, d_{480}\}$: A corpus of legal documents, where $d_j$ denotes a single legal document (e.g., statutes, administrative regulations).
    \item $\text{Splitter}: D \to C \cup A$: A document splitting function that divides $d_j$ into a \textit{chunk set} $C$ (Chunk-based) and an \textit{article set} $A$ (Article-based), corresponding to the dual-path splitting of the "Splitter" module in the framework diagram.
    \item $\text{Index}: C \cup A \to DB$: An index construction function that converts $C/A$ into vector/keyword indexes via "Vectorization (embedding)" and "Retriever create (Keyword)", which are stored in the Database—matching the input process of the Database in the framework diagram.
    \item $\text{1st Retrieval}: q \times DB \to C_{\text{top20}}$: The first-stage retrieval that recalls the top 20 relevant chunks from the Database based on "Embedding + Keywords", corresponding to the "1st Retrieval" module in the framework diagram.
    \item $\text{2nd Retrieval}: C_{\text{top20}} \to C_{\text{top}K}$: The second-stage reranking that filters the top $K$ relevant chunks (with $K=5$ in this study) via "Rerank", corresponding to the "2nd Retrieval" module in the framework diagram.
    \item $\text{Gen}: q \times C_{\text{top}K} \to \hat{a}_{\text{init}}$: An initial generation function that produces the initial answer $\hat{a}_{\text{init}}$ based on "Prompt \& LLM".
    \item $\text{Self-Reflection}: q \times \hat{a}_{\text{init}} \times C_{\text{top}K} \to \hat{a}$: A self-verification function that checks the legal accuracy and completeness of $\hat{a}_{\text{init}}$ against $C_{\text{top}K}$ under query context $q$, and outputs the final answer $\hat{a}$.
\end{itemize}

The core objective of the framework is to maximize the similarity between $\hat{a}$ and the gold reference answer $a^*$ through the full-process optimization of "Splitting $\to$ Dual Retrieval $\to$ Generation + Self-Verification", as formally expressed in \Cref{eq:legrad_formal}.

\clearpage

\begin{strip}
\begin{equation}
\begin{aligned}
\hat{a} = \text{Self-Reflection}\Big(
&q,\ \text{Gen}\big(
q,\ \text{2nd Retrieval}\big( \\
&\text{1st Retrieval}\big(
q,\ \text{Index}(\text{Splitter}(D))
\big)
\big)
\big)
\Big)
\end{aligned}
\label{eq:legrad_formal}
\tag{3}
\end{equation}
\end{strip}

\subsection{Modular Implementation of LegRAG}
The LegRAG framework (\Cref{fig:legrag_architecture}) consists of six core modules. The function and data flow of each module are detailed as follows:

\subsubsection{Splitter Module}
Corresponding to the "Splitter" node in the framework diagram, this module is responsible for splitting original legal documents $D$ into two granularities of content:
\begin{itemize}
    \item \textit{Chunk-based Splitting}: Splits documents by semantic paragraphs to retain contextual details (e.g., "application scenarios" of a clause).
    \item \textit{Article-based Splitting}: Splits documents by legal article numbers (e.g., treating "Article 19" as an independent unit) to ensure the integrity of legal provisions.
    \item \textit{Output}: A chunk set $C$ and an article set $A$, matching the dual-path output of "token flow" and "doc flow" in the framework diagram.
\end{itemize}

\subsubsection{Indexing Module}
Corresponding to the "Vectorization (embedding)" and "Retriever create (Keyword)" nodes in the framework diagram, this module converts $C/A$ into retrievable indexes:
\begin{itemize}
    \item \textit{Vector Index}: Converts $C/A$ into vectors using an embedding model (e.g., ColBERT-Legal) and stores them in the Database to support "semantic similarity retrieval".
    \item \textit{Keyword Index}: Extracts legal keywords (e.g., "real-name registration", "administrative sanction") from $C/A$ to build a keyword index, supporting "exact matching retrieval".
    \item \textit{Output}: A hybrid index database integrating vectors and keywords.
\end{itemize}

\subsubsection{1st Retrieval Module}
Corresponding to the "1st Retrieval (Embedding \& Keywords)" node in the framework diagram, this module takes the query $q$ and Database as inputs, and recalls the top 20 relevant chunks $C_{\text{top20}}$ via a hybrid strategy of "vector retrieval + keyword retrieval"—matching the output of "chunks \#1-\#20" in the framework diagram.

\subsubsection{2nd Retrieval Module}
Corresponding to the "2nd Retrieval (Rerank)" node in the framework diagram, this module takes $C_{\text{top20}}$ as input and filters the top 5 relevant chunks $C_{\text{top5}}$ using a reranking model (e.g., Qwen3-Rerank)—matching the output of "reranked chunks \#6/\#4/\#17" in the framework diagram.

\subsubsection{Prompt \& LLM Module}
Corresponding to the "Prompt \& LLM" node in the framework diagram, this module takes $q$ and $C_{\text{top5}}$ as inputs, and guides the LLM to generate the initial answer $\hat{a}_{\text{init}}$ using legal-specific prompts. The prompt logic follows the principle of "first selecting relevant chunks from $C_{\text{top5}}$, then generating answers based on the chunks" (as shown in \Cref{fig:prompt_answer_generation}).

\subsubsection{Self-Reflection \& Verification Module}
Corresponding to the "Self-Reflection \& Verification" node in the framework diagram, this module’s core function is \textit{closed-loop verification}:
\begin{itemize}
    \item \textit{Verification Dimensions}:
    \begin{enumerate}[label=(\arabic*),noitemsep,topsep=0pt,leftmargin=*]
        \item Whether $\hat{a}_{\text{init}}$ is fully based on $C_{\text{top5}}$ (no external knowledge).
        \item Whether legal conclusions are consistent with provisions in $C_{\text{top5}}$.
        \item Whether key information in $C_{\text{top5}}$ is omitted.
    \end{enumerate}
    \item \textit{Output}: If verification passes, $\hat{a}$ is directly output; if not, the module returns to the "2nd Retrieval" module for supplementary retrieval or to the "Prompt \& LLM" module for re-generation—forming a closed-loop optimization.
\end{itemize}

\subsection{Experimental Setup}
To verify the effectiveness of the updated LegRAG framework, three mainstream RAG systems were selected for comparative experiments, with settings aligned with the framework modules:
\begin{itemize}[leftmargin=*]
    \item \textit{Evaluation Metrics}: For the retrieval module: Recall@20, MRR@5; For the generation module: Accuracy, ROUGE-L, LLM-Score (Qwen3-max as the automated evaluator).
    \item \textit{Experimental Data}: Legal-DC dataset (480 documents, 2475 QA pairs), with the retrieval candidate size set to $K=5$.
\end{itemize}

By guiding LLMs with a highly capable and efficient prompting strategy, there is significant room for improvement in boosting specialist performance \cite{18}. Our prompt settings are shown in \Cref{fig:prompt_answer_generation}. We steer LLMs to select the correct documents and provide answers using prompts. We selected QAnything, LightRAG and Weknora for comparative experiments on our benchmark.

\begin{table}[t]
    \centering
    \caption{Retrieval Results (\%) of Different Methods.}
    \label{tab:retrieval}
    \begin{tabular*}{\columnwidth}{@{\extracolsep{\fill}}lcc@{}}
    \toprule
    Method    & Recall & MRR@5  \\
    \midrule
    Weknora   & 61.35  & 35.44  \\
    QAnything & 63.69  & 36.67  \\
    LightRAG  & 75.86  & 47.56  \\
    LegRAG    & \textbf{78.02}  & \textbf{50.23} \\
    \bottomrule
    \end{tabular*}
\end{table}

\begin{table}[t]
    \centering
    \caption{Precision Comparison of Different Models}
    \label{tab:model_precision}
    \small
    \begin{tabular*}{\columnwidth}{@{\extracolsep{\fill}}l c@{}}
        \toprule
        Model & Precision(\%) \\
        \midrule
        Qwen3-max & \textbf{45.32} \\
        DeepSeek-V3.2 & 40.23 \\
        \bottomrule
    \end{tabular*}
\end{table}

\section{Results and Analysis}

\subsection{Retrieval Results}

\Cref{tab:retrieval} reports the retrieval performance of different methods on Legal-DC. LegRAG achieves the best results on both metrics, with Recall = 78.02\% and MRR@5 = 50.23\%. Compared with the strongest baseline (LightRAG), LegRAG improves Recall by 2.16 percentage points and MRR@5 by 2.67 percentage points, indicating better ranking quality for relevant legal content.

\begin{table*}[t]
\centering
\caption{Performance Comparison of Different RAG Frameworks with Various LLMs}
\label{tab:rag_framework_comparison}
\renewcommand{\arraystretch}{1.1}
\small

\resizebox{\textwidth}{!}{
\begin{tabular}{llcccccccc}
\toprule
\textbf{Framework} & \textbf{Model} & \textbf{Recall (\%)} & \textbf{MRR@5 (\%)} & \textbf{Precision (\%)} & \textbf{BLEU (\%)} & \textbf{ROUGE (\%)} & \textbf{MAP (\%)} & \textbf{F1 (\%)} & \textbf{LLM-eval (\%)} \\
\midrule

\multirow{3}{*}{Weknora}
 & DeepSeek-V3.2 & 57.47 & 32.74 & 50.98 & 38.43 & 41.42 & 30.59 & 55.86 & 60.82 \\
 & Qwen3-8B      & 59.36 & 33.44 & 51.86 & 37.49 & 38.89 & 31.84 & 57.08 & 62.71 \\
 & Qwen3-max     & 61.35 & 35.44 & 52.28 & 38.61 & 42.59 & 33.49 & 57.48 & 64.68 \\
\addlinespace

\multirow{3}{*}{QAnything}
 & DeepSeek-V3.2 & 58.61 & 32.33 & 50.19 & 36.18 & 42.78 & 30.46 & 55.43 & 62.03 \\
 & Qwen3-8B      & 59.78 & 34.01 & 51.21 & 36.67 & 44.15 & 31.69 & 55.05 & 63.15 \\
 & Qwen3-max     & 63.69 & 36.67 & 52.58 & 40.93 & 43.34 & 34.70 & 58.99 & 67.02 \\
\addlinespace

\multirow{3}{*}{LightRAG}
 & DeepSeek-V3.2 & 75.54 & 44.99 & 56.16 & 46.43 & 45.79 & 37.83 & 64.06 & 78.74 \\
 & Qwen3-8B      & 75.62 & 46.50 & 56.93 & 46.34 & 50.47 & 38.99 & 65.86 & 78.81 \\
 & Qwen3-max     & 75.86 & 47.56 & 58.58 & 46.92 & 48.41 & 38.58 & 64.99 & 79.05 \\
\addlinespace

\multirow{3}{*}{Legal-DC}
 & DeepSeek-V3.2 & 74.51 & 44.43 & 57.91 & 45.27 & 48.12 & 38.23 & 64.34 & 77.86 \\
 & Qwen3-8B      & 77.01 & 45.77 & 58.80 & 43.84 & 46.43 & 38.97 & 65.62 & 80.13 \\
 & Qwen3-max     
 & \textbf{78.02} {\scriptsize(+2.16)}
 & \textbf{50.23} {\scriptsize(+2.67)}
 & \textbf{59.56} {\scriptsize(+0.98)}
 & 46.68
 & \textbf{51.27} {\scriptsize(+0.80)}
 & \textbf{42.21} {\scriptsize(+3.22)}
 & 65.76
 & \textbf{81.25} {\scriptsize(+2.20)}\\

\bottomrule
\end{tabular}
}

\end{table*}

\begin{table*}[t]
    \centering
    \caption{Performance Comparison on Conceptual, Generalization, and Logic Task Dimensions}
    \label{tab:task_dimension_performance}
    \renewcommand{\arraystretch}{1.15}
    \small
    \resizebox{\textwidth}{!}{
    \begin{tabular}{llccccccccc}
    \toprule
    \textbf{Framework} & \textbf{Model} 
    & \multicolumn{3}{c}{\textbf{Conceptual}} 
    & \multicolumn{3}{c}{\textbf{Generalization}} 
    & \multicolumn{3}{c}{\textbf{Logic}} \\
    \cmidrule(lr){3-5} \cmidrule(lr){6-8} \cmidrule(lr){9-11}
                      &                
                      & Acc (\%) & BLEU (\%) & ROUGE (\%)  
                      & Acc (\%) & BLEU (\%) & ROUGE (\%)  
                      & Acc (\%) & BLEU (\%) & ROUGE (\%)  \\
    \midrule
    \multirow{3}{*}{Weknora}
        & DeepSeek-V3.2 & 57.86 & 37.96 & 39.96 & 59.04 & 37.48 & 41.37 & 57.57 & 38.51 & 40.87 \\
        & Qwen3-8B      & 59.68 & 38.23 & 42.70 & 58.93 & 39.46 & 40.04 & 59.49 & 38.21 & 43.19 \\
        & Qwen3-max     & 61.62 & 41.12 & 42.79 & 62.71 & 38.58 & 43.29 & 62.72 & 39.52 & 41.29 \\
    \addlinespace

    \multirow{3}{*}{QAnything}
        & DeepSeek-V3.2 & 60.02 & 40.41 & 42.86 & 61.39 & 38.89 & 40.20 & 60.32 & 38.64 & 42.08 \\
        & Qwen3-8B      & 61.36 & 40.29 & 43.13 & 59.93 & 38.61 & 43.98 & 60.94 & 38.19 & 40.78 \\
        & Qwen3-max     & 62.19 & 41.14 & 43.27 & 62.89 & 40.86 & 44.38 & 61.97 & 39.05 & 43.16 \\
    \addlinespace

    \multirow{3}{*}{LightRAG}
        & DeepSeek-V3.2 & 74.05 & 46.56 & 48.21 & 70.15 & 46.12 & 47.83 & 73.91 & 44.42 & 46.24 \\
        & Qwen3-8B      & 76.32 & 46.58 & 49.37 & 70.59 & 47.02 & 47.31 & 74.76 & 45.29 & 47.56 \\
        & Qwen3-max     & 75.71 & 47.73 & 49.73 & 71.38 & 46.58 & 48.17 & 76.65 & 46.24 & 48.03 \\
    \addlinespace

    \multirow{3}{*}{Legal-DC}
        & DeepSeek-V3.2 & 76.72 & 45.78 & 49.32 & 74.21 & 46.48 & 47.02 & 75.83 & 46.36 & 48.20 \\
        & Qwen3-8B      & 76.01 & 46.17 & 48.39 & 75.78 & 46.72 & 49.24 & 75.33 & 47.83 & 49.18 \\
        & Qwen3-max     
        & \textbf{78.23} {\scriptsize (+1.91)}
        & 47.43
        & 49.52
        & \textbf{77.03} {\scriptsize (+5.65)}
        & \textbf{48.32} {\scriptsize (+1.30)}
        & \textbf{51.23} {\scriptsize (+3.06)}
        & 76.12
        & \textbf{46.67} {\scriptsize (+0.43)}
        & \textbf{53.42} {\scriptsize (+5.39)} \\
    \bottomrule
    \end{tabular}
    }
\end{table*}

\FloatBarrier

\subsection{Generation Evaluation Results}

\Cref{tab:model_precision} compares generation precision without retrieval augmentation. Both models show limited standalone performance (Qwen3-max: 0.4532; DeepSeek-V3.2: 0.4023), confirming that external evidence retrieval is necessary in legal consultation scenarios.

\Cref{tab:rag_framework_comparison} shows full-pipeline results across frameworks and LLMs. The best overall setting is LegRAG + Qwen3-max, which achieves the highest values on Recall, MRR@5, Precision, F1, and LLM-eval among all compared combinations.

\Cref{tab:task_dimension_performance} further breaks down performance by Conceptual, Generalization, and Logic dimensions. Legal-DC remains consistently stronger than LightRAG, QAnything, and Weknora across most settings, while stronger backbone models (especially Qwen3-max) provide additional gains.

\Cref{tab:processing_strategy_comparison} demonstrates that combining chunk-based and article-based processing provides the strongest results compared with either single strategy. \Cref{tab:entity_ablation} and \Cref{tab:self_reflection} show that entity augmentation and self-reflection both improve retrieval and generation metrics, with self-reflection yielding large overall gains (e.g., for Qwen3-max, F1 improves from 0.6576 to 0.6928).

\Cref{tab:question_type_performance} compares models across question types under the same setting. Qwen3-max ranks first on all types and overall, while DeepSeek-V3.2 is the weakest.

\begin{table*}[t]
\centering
\caption{Performance Comparison of Different Processing Strategies under Various LLMs}
\label{tab:processing_strategy_comparison}
\renewcommand{\arraystretch}{1.1}
\small
\resizebox{\textwidth}{!}{
\begin{tabular}{llcccccccc}
\toprule
\textbf{Processing Strategy} & \textbf{Model} 
& Recall (\%) & MRR@5 (\%) & Precision (\%) & BLEU (\%) & ROUGE (\%) & MAP (\%) & F1 (\%) & LLM-eval (\%) \\
\midrule
\addlinespace
\multirow{3}{*}{Chunk-based}
 & Qwen3-8B      & 74.51 & 44.43 & 57.91 & 48.96 & 48.12 & 38.23 & 64.34 & 67.25 \\
 & DeepSeek-V3.2 & 77.01 & 45.77 & 58.80 & 43.84 & 46.43 & 38.97 & 65.62 & 63.19 \\
 & Qwen3-max     & 78.02 & 48.23 & 59.56 & 46.68 & 51.27 & 42.21 & 65.76 & 62.54 \\
\addlinespace

\multirow{3}{*}{Article-based}
 & Qwen3-8B      & 76.21 & 45.21 & 58.21 & 49.12 & 49.66 & 39.25 & 66.89 & 75.02 \\
 & DeepSeek-V3.2 & 80.43 & 47.33 & 60.27 & 45.23 & 48.65 & 40.87 & 66.45 & 64.68 \\
 & Qwen3-max     & 82.12 & 50.32 & 62.45 & 52.12 & 56.23 & 46.23 & 67.34 & 66.15 \\
\addlinespace

\multirow{3}{*}{Chunk + Article}
 & Qwen3-8B      & 77.45 & 46.45 & 59.80 & 50.67 & 51.65 & 41.72 & 67.65 & 83.21 \\
 & DeepSeek-V3.2 & 82.34 & 48.67 & 61.34 & 46.34 & 49.43 & 42.06 & 67.43 & 80.81 \\
 & Qwen3-max     
 & \textbf{85.23} {\scriptsize (+3.11)}
 & \textbf{51.43} {\scriptsize (+1.11)}
 & \textbf{63.67} {\scriptsize (+1.22)}
 & \textbf{54.52} {\scriptsize (+2.40)}
 & \textbf{57.56} {\scriptsize (+1.33)}
 & \textbf{48.76} {\scriptsize (+2.53)}
 & \textbf{68.30} {\scriptsize (+0.96)}
 & \textbf{78.74} {\scriptsize (+3.72)}\\
\bottomrule
\end{tabular}
}
\end{table*}

\begin{table*}[t]
\centering
\caption{Effect of Entity Augmentation on Retrieval and Generation Performance}
\label{tab:entity_ablation}
\renewcommand{\arraystretch}{1.1}
\small
\resizebox{\textwidth}{!}{
\begin{tabular}{llccccccc}
\toprule
\textbf{Entity Augmentation} & \textbf{Model} & Recall (\%) & MRR@5 (\%) & Precision (\%) & BLEU (\%) & ROUGE (\%) & MAP (\%) & F1 (\%) \\
\midrule
\addlinespace
\multirow{3}{*}{NO}
 & DeepSeek-V3.2  & 73.52 & 43.23 & 57.63 & 47.86 & 47.13 & 37.25 & 63.22 \\
 & Qwen3-8B       & 76.70 & 44.56 & 57.43 & 46.47 & 45.34 & 38.09 & 64.52 \\
 & Qwen3-max      & 77.12 & 46.63 & 58.54 & 46.34 & 50.12 & 43.52 & 64.35 \\
\addlinespace
\multirow{3}{*}{Yes}
 & DeepSeek-V3.2  & 75.32 & 46.25 & 57.34 & 50.56 & 49.27 & 38.62 & 65.48 \\
 & Qwen3-8B       & 81.44 & 47.25 & 61.76 & 46.33 & 48.34 & 41.89 & 65.44 \\
 & Qwen3-max      
 & \textbf{82.73}{\scriptsize (+5.61)}
 & \textbf{51.34}{\scriptsize (+4.71)}
 & \textbf{62.90}{\scriptsize (+4.36)}
 & \textbf{53.15}{\scriptsize (+5.29)}
 & \textbf{54.26}{\scriptsize (+4.14)}
 & \textbf{45.25}{\scriptsize (+1.73)}
 & \textbf{66.24}{\scriptsize (+1.72)} \\
\bottomrule
\end{tabular}
}
\end{table*}

\begin{table*}[t]
\centering
\caption{Performance Comparison of Legal-DC Systems With and Without Self-Reflection}
\label{tab:self_reflection}
\renewcommand{\arraystretch}{1.1}
\small
\resizebox{\textwidth}{!}{
\begin{tabular}{llccccccc}
\toprule
\textbf{Reflection} & \textbf{Model} & Recall (\%) & MRR@5 (\%) & Precision (\%) & BLEU (\%) & ROUGE (\%) & MAP (\%) & F1 (\%) \\
\midrule
\multirow{3}{*}{No}
 & DeepSeek-V3.2 & 74.51 & 44.43 & 57.91 & 48.96 & 48.12 & 38.23 & 64.34 \\
 & Qwen3-8B      & 77.01 & 45.77 & 58.80 & 43.84 & 46.43 & 38.97 & 65.62 \\
 & Qwen3-max     & 78.02 & 48.23 & 59.56 & 46.68 & 51.27 & 42.21 & 65.76 \\
\addlinespace
\multirow{3}{*}{Yes}
 & DeepSeek-V3.2 & 79.65 & 48.20 & 61.24 & 52.49 & 52.47 & 42.65 & 68.26 \\
 & Qwen3-8B      & 84.21 & 49.78 & 62.44 & 48.12 & 51.24 & 44.19 & 69.44 \\
 & Qwen3-max     
 & \textbf{86.34}{\scriptsize (+2.13)}
 & \textbf{53.54}{\scriptsize (+5.31)}
 & \textbf{65.60}{\scriptsize (+6.04)}
 & \textbf{57.23}{\scriptsize (+8.27)}
 & \textbf{59.11}{\scriptsize (+7.84)}
 & \textbf{50.32}{\scriptsize (+8.11)}
 & \textbf{69.28}{\scriptsize (+3.52)} \\
\bottomrule
\end{tabular}
}
\end{table*}

\begin{table*}[!tbp]
    \centering
    \caption{Performance Comparison Across Different Question Types}
    \label{tab:question_type_performance}
    \small
    \begin{tabular*}{\textwidth}{@{\extracolsep{\fill}}lcccc@{}}
        \toprule
        \textbf{Model} 
        & \textbf{Conceptual (\%)} 
        & \textbf{Generalization (\%)} 
        & \textbf{Logic (\%)} 
        & \textbf{Avg (\%)} \\
        \midrule
        DeepSeek-V3.2 & 69.90 & 66.10 & 62.80 & 66.27 \\
        Qwen3-8B      & 73.00 & 69.50 & 66.30 & 69.60 \\
        Qwen3-max     & \textbf{76.00} & \textbf{72.50} & \textbf{69.60} & \textbf{72.70} \\
        \bottomrule
    \end{tabular*}
\end{table*}

\subsubsection{Difficulty Analysis Across Question Types}

A systematic analysis of performance across question types reveals a clear difficulty gradient. As shown in \Cref{tab:question_type_performance}, all three models achieve the highest average score on Conceptual questions (0.730), followed by Generalization (0.694), while Logic is the most difficult (0.662).

The difficulty differences stem from the distinct cognitive and retrieval requirements of each question type. Conceptual questions primarily rely on precise keyword matching and definition extraction from legal provisions (e.g., "What is real-name registration?"), which aligns well with the streng-ths of embedding-based retrieval. In contrast, Generalization questions require synthesizing information across multiple clauses and abstracting key points, demanding higher-level semantic understanding. Logic questions pose the greatest challenge as they necessitate capturing implicit inferential relationships between provisions (e.g., "Does purchasing a drone trigger registration requirements?" requires linking 'purchase' → 'ownership' → 'registration obligation'), which are rarely expressed through explicit keywords.

This pattern is also reflected in dimension-level results in \Cref{tab:task_dimension_performance}. For Legal-DC + Qwen3-max, Accuracy decreases from 0.7823 (Conceptual) to 0.7612 (Logic), and BLEU decreases from 0.4743 to 0.4667. ROUGE shows a different trend (0.4952 to 0.5342), indicating that lexical overlap alone does not fully capture reasoning difficulty in legal QA.

These findings support the use of dual-path retrieval and self-reflection in LegRAG. As shown in \Cref{tab:self_reflection}, self-reflection improves all metrics across all tested models, demonstrating robust benefits for legal answer verification and refinement.

\subsection{Case Study}

We analyzed representative failure cases (\Cref{tab:food_safety_responsibility,tab:reasoning_retrieval_failure_2,tab:reasoning_retrieval_failure}) and summarize the main causes below. Detailed case tables are provided in Appendix A.

For retrieval failures:

\begin{enumerate}[label=(\arabic*), noitemsep, topsep=0pt, leftmargin=*]
    \item \textbf{Keyword frequency bias.} The retriever may prefer passages with frequent surface keywords over legally relevant passages. In \Cref{tab:food_safety_responsibility}, retrieved evidence over-focuses on ``responsibility'' wording while missing the authoritative clause that actually answers the question.
    \item \textbf{Weak support for implicit logical links.} For reasoning-type questions, lexical overlap is often insufficient. In \Cref{tab:reasoning_retrieval_failure_2}, retrieval over-matches ``purchase''-related content but misses the provision connecting ownership status to real-name registration duty.
    \item \textbf{Insufficient conjunctive matching.} Questions requiring multiple constraints are vulnerable when retrieval satisfies only part of the conditions.
    \item \textbf{Fragmented matching for long legal phrases.} Long legal terms are sometimes partially matched, reducing retrieval precision.
\end{enumerate}

For response failures:

\begin{enumerate}[label=(\arabic*), noitemsep, topsep=0pt, leftmargin=*]
    \item \textbf{Answering without valid evidence.} When relevant evidence is not retrieved, the generator may still produce fluent but weakly grounded answers based on irrelevant context.
    \item \textbf{Suboptimal document selection under mixed evidence.} Even when the correct clause is present in the candidate set, the model may prioritize a narrower but less appropriate clause. \Cref{tab:reasoning_retrieval_failure} shows this pattern in contract-template authority identification.
\end{enumerate}

\section{Conclusions and Future Work}
\subsection{Key Findings and Contributions}
This study addresses the joint evaluation and optimization problem of Chinese legal RAG systems through benchmark construction, framework design, and systematic experiments.

\textbf{Benchmark contribution.}
We built Legal-DC, a benchmark containing 480 legal documents and 2,475 high-quality QA pairs with fine-grained reference annotations. The dataset supports joint evaluation of retrieval relevance and answer faithfulness, with strong annotation consistency (Cohen's Kapp
a $\ge$ 0.83).

\textbf{Method contribution.}
We proposed LegRAG, formally defined in \Cref{eq:legrad_formal}, with legal-adaptive indexing, dual-path retrieval, and self-reflection verification. In retrieval, LegRAG outperforms strong baselines (LightRAG) on both Recall (78.02\% vs. 75.86\%) and MRR@5 (50.23\% vs. 47.56\%) as shown in \Cref{tab:retrieval}.

\textbf{Empirical findings.}
Experiments show that retrieval granularity is a key factor in legal RAG performance, and that combining chunk-level and article-level processing improves overall effectiveness. Ablation results further confirm that entity augmentation and self-reflection consistently improve retrieval-generation alignment and answer quality.

\subsection{Limitations}
Although the proposed framework achieves consistent improvements, two limitations remain.
\begin{enumerate}[label=(\arabic*),noitemsep,topsep=0pt,leftmargin=*]
    \item \textbf{Reasoning difficulty}: Logic-intensive legal questions remain the most challenging type, indicating that current retrieval and generation modules still have limited capability in multi-hop legal inference.
    \item \textbf{Evaluation dependence}: The automatic evaluation protocol is highly correlated with human judgments, but still relies on an LLM evaluator and may not fully replace large-scale expert review in high-stakes legal settings.
\end{enumerate}

\vspace{-0.35\baselineskip}
\subsection{Future Work}
Based on the current framework and dataset, we plan to explore three directions:
\begin{enumerate}[label=(\arabic*),noitemsep,topsep=0pt,leftmargin=*]
    \item \textbf{Enhance Logical Reasoning for Legal Queries}: To address the reasoning bottleneck, we will introduce a legal logic rule base to supplement retrieval and generation, enabling better modeling of implicit clause dependencies (e.g., condition-to-obligation chains).
    \item \textbf{Strengthen Evaluation Reliability}: To reduce dependence on a single automatic evaluator, we will build a hybrid protocol that combines LLM-based scoring with periodic expert calibration and disagreement analysis.
    \item \textbf{Open-Source the LegRAG Toolkit}: We will package the framework into a reproducible toolkit, including legal embeddings, clause-level chunking utilities, and self-reflection prompts, to support broader adoption and standardized comparison.
\end{enumerate}

\fontsize{8.5pt}{10pt}\selectfont

\appendix
\setcounter{secnumdepth}{0}
\section{Acknowledgements}
This research was supported by the National Natural Science Foundation
of China (NSFC) under Grant Nos. 62102044 and 72501034.

\section{Competing interests}
The authors declare that they have no competing interests or financial conflicts to disclose.

\balance
\bibliographystyle{plain}
\bibliography{ref}

\section{Appendix A. Detailed Case Study Tables}

\setcounter{table}{0}
\renewcommand{\thetable}{A\arabic{table}}

\begin{table*}[t]
\centering
\caption{Data Samples}
\label{tab:data_samples}
\scriptsize
\setlength{\tabcolsep}{2pt}
\renewcommand{\arraystretch}{0.92}

\begin{tabularx}{\textwidth}{p{3.2cm} X X p{1.6cm}}
\toprule
\textbf{Question} & \textbf{Answer} & \textbf{Reference content} & \textbf{Type} \\
\midrule

外国投资者或者外商投资企业发现其存在未报、错报、漏报有关投资信息的应该怎么办？  
& 外国投资者或者外商投资企业发现其存在未报、错报、漏报有关投资信息的，应当及时进行补报或更正。商务主管部门发现相关情形的，应当通知其于20个工作日内补报或更正；涉及公示事项的，更正前后的信息应当同时公示。  
& 第十九条 外国投资者或者外商投资企业发现其存在未报、错报、漏报有关投资信息的，应当及时进行补报或更正。商务主管部门发现相关情形的，应当通知其于 20 个工作日内补报或更正；涉及公示事项的，更正前后的信息应当同时公示。  
& Concept explanation \\

(What should foreign investors or foreign-invested enterprises do if they discover that they have failed to report, misreported, or omitted relevant investment information?)  
& (Foreign investors or foreign-invested enterprises that discover any unreported, misreported, or omitted investment information shall promptly submit supplementary reports or corrections……)
& (Article 19 Where a foreign investor or a foreign-invested enterprise discovers that it has failed to report, misreported, or omitted relevant investment information, it shall promptly make supplementary reports or corrections…………)  
& \\

\midrule

某外国投资者拟设立外资企业，计划认缴注册资本为 800 万美元，且以工业产权作价出资。同时，该企业的投资总额拟定为 2000 万美元。 该企业的注册资本是否符合 “注册资本与投资总额比例” 的法定要求?     
& 符合规定。该企业投资总额为 2000 万美元，属于 “1000 万美元以上至 3000 万美元” 的区间，法定注册资本不得低于投资总额的 2/5。计算：2000万美元×2/5=800万美元，而该企业拟定的注册资本为 800 万美元，恰好满足法定最低要求，因此符合规定。  
& 第三条投资总额在 1000 万美元以上至 3000 万美元（含 3000 万美元）的，其注册资本至少应占投资总额的 2/5；其中投资总额在 1250 万美元以下的，注册资本不得低于 500 万美元。
& Logical reasoning \\

(A foreign investor intends to establish a foreign-invested enterprise with a planned subscribed registered capital of US\$8 million, to be contributed in the form of industrial property rights. Concurrently……)   
& (Compliant with regulations. The enterprise's total investment amount is
US\$20 million, falling within the range of "US\$10 million to \$30 million" range, where the statutory registered capital must not be less than 2/5 of the total investment. Calculation: \$20 million × 2/5 = \$8 million. The enterprise's proposed registered capital of \$8 million precisely meets the statutory minimum requirement and is therefore compliant.)  
& (For investments totaling between US\$10 million and US\$30 million (inclusive), the registered capital shall constitute at least two-fifths of the total investment amount. Where the total investment amount is below US\$12.5 million, the registered capital shall not be less than US\$5 million.)  
& \\

\midrule

违反《安全技术防范产品管理办法》的规定，相关工作人员滥用职权、玩忽职守的，将面临怎样的处罚？  
& 由有关部门按照干部管理权限给予行政处分；构成犯罪的，依法追究刑事责任。  
& 第十七条 滥用职权、玩忽职守的，依法给予处分；构成犯罪的，依法追究刑事责任。  
& Generalization and Synthesis \\

(What penalties will staff members face for abuse of authority or dereliction of duty?)  
& (They shall be subject to administrative sanctions; criminal liability shall be pursued where a crime is constituted.)  
& (Article 17 Staff members who abuse authority or neglect duties shall be punished according to law.)  
& \\

\bottomrule
\end{tabularx}
\end{table*}

\begin{table*}[!tbp]
\centering
\caption{Information about Responsibility for Food Safety Sampling and Inspection}
\label{tab:food_safety_responsibility}
\fontsize{6.8pt}{7.6pt}\selectfont
\setlength{\tabcolsep}{2pt}
\renewcommand{\arraystretch}{0.88}
\begin{tabularx}{\textwidth}{p{0.09\textwidth} X}
    \toprule
    \textbf{Query} &
    谁负责食品安全抽样检验工作？(Who is \textbf{responsible} for food safety sampling and inspection?) \\
    \midrule
    \textbf{Answer} &
    食品安全抽样检验工作由国家市场监督管理总局和县级以上地方市场监督管理部门协同负责。
    (Food safety sampling and inspection are jointly managed by the State Administration for Market Regulation and local market supervision departments at or above the county level.) \\
    \midrule
    \textbf{Gold Content} &
    第六条国家市场监督管理总局建立国家食品安全抽样检验信息系统，定期分析食品安全抽样检验数据，
    加强食品安全风险预警，完善并督促落实相关监督管理制度。县级以上地方市场监督管理部门应当按照规定
    通过国家食品安全抽样检验信息系统，及时报送并汇总分析食品安全抽样检验数据。
    (Article 6: The State Administration for Market Regulation establishes a national food safety sampling and inspection information system, periodically analyzes food safety sampling and inspection data, strengthens food safety risk warnings, and improves and supervises the implementation of relevant supervision and management systems. Local market supervision departments at or above the county level shall, in accordance with regulations, promptly submit and summarize food safety sampling and inspection data through the national food safety sampling and inspection information system.) \\
    \midrule
    \textbf{Retrieved Content} &
    第二十四条食品安全抽样检验实行承检机构与检验人负责制。承检机构出具的食品安全检验报告应当加盖机构公章，
    并有检验人的签名或者盖章。承检机构和检验人对出具的食品安全检验报告负责。承检机构应当自收到样品之日起
    20个工作日内出具检验报告。市场监督管理部门与承检机构另有约定的，从其约定。未经组织实施抽样检验任务的
    市场监督管理部门同意，承检机构不得分包或者转包检验任务。
    (Article 24: Food safety sampling and inspection follow a \textbf{responsibility} system for inspection agencies and inspectors. Inspection agencies and inspectors are \textbf{responsible} for issued inspection reports.) \\
    & 第三、四、七、八条略。(Articles 3, 4, 7, 8: omitted.) \\
    \midrule
    \textbf{LLM Answer} &
    食品安全抽样检验工作由国家市场监督管理总局和县级以上地方市场监督管理部门协同负责。
    (Food safety sampling and inspection are jointly managed by the State Administration for Market Regulation and local market supervision departments at or above the county level.) \\
    \bottomrule
\end{tabularx}
\end{table*}

\begin{table*}[!tbp]
\centering
\caption{Example of Logical Reasoning Retrieval Failure}
\label{tab:reasoning_retrieval_failure_2}
\fontsize{6.8pt}{7.6pt}\selectfont
\setlength{\tabcolsep}{2pt}
\renewcommand{\arraystretch}{0.88}
\begin{tabularx}{\textwidth}{p{0.09\textwidth} X}
    \toprule
    \textbf{Query} &
    购买微型无人机必须实名登记吗？(Is real-name registration required for purchasing micro drones?) \\
    \midrule
    \textbf{Answer} &
    是的，民用无人驾驶航空器所有者应当依法进行实名登记。
    (Yes, owners of civil unmanned aerial vehicles shall legally perform real-name registration.) \\
    \midrule
    \textbf{Gold Content} &
    第十条民用无人驾驶航空器所有者应当依法进行实名登记，具体办法由国务院民用航空主管部门会同有关部门制定。
    (Article 10: Owners of civil unmanned aerial vehicles shall legally perform real-name registration, with specific measures formulated by the civil aviation authority under the State Council in conjunction with relevant departments.) \\
    \midrule
    \textbf{Retrieved Content} &
    第十七条购买第二类、第三类易制毒化学品的，应当在购买前将所需购买的品种、数量，
    向所在地的县级人民政府公安机关备案。个人自用购买少量高锰酸钾的，无须备案。
    (Article 17: For the purchase of second or third category precursor chemicals, the variety and quantity to be purchased shall be filed with the local public security organ at the county level or above before purchase.) \\
    & 第十二、七十八、四十三、八十七条略。(Articles 12, 78, 43, 87: omitted.) \\
    \midrule
    \textbf{LLM Answer} &
    民用无人驾驶航空器所有者应当依法进行实名登记。
    (Owners of civil unmanned aerial vehicles shall legally perform real-name registration.) \\
    \bottomrule
\end{tabularx}
\end{table*}

\begin{table*}[!t]
    \centering
    \caption{Example of Response Failure Due to Incorrect Document Selection}
    \label{tab:reasoning_retrieval_failure}
    \footnotesize
    \renewcommand{\arraystretch}{1.05}
    \begin{tabularx}{\textwidth}{p{0.15\textwidth} X}
        \toprule
    Query & 合同示范文本由谁负责制定？(Who is responsible for formulating contract template texts?) \\
    \midrule
    Answer & 省级以上市场监督管理部门可以根据有关法律法规规定，针对特定行业或者领域，联合有关部门制定合同示范文本。(Market supervision and administration departments at or above the provincial level may, in accordance with relevant laws and regulations, jointly formulate contract template texts for specific industries or fields with relevant departments.) \\
    \midrule
    Gold Content & 第十三条省级以上市场监督管理部门可以根据有关法律法规规定，针对特定行业或者领域，联合有关部门制定合同示范文本。(Article 13: Market supervision and administration departments at or above the provincial level may, in accordance with relevant laws and regulations, jointly formulate contract template texts for specific industries or fields with relevant departments.) \\
    \midrule
    Retrieved Content & 第十六条进行二手车交易应当签订合同。合同示范文本由国务院工商行政管理部门制定。(Article 16: A contract shall be signed for second-hand car transactions. The contract template text is formulated by the State Council's administrative department for industry and commerce.) \\
    & 第十三条省级以上市场监督管理部门可以根据有关法律法规规定，针对特定行业或者领域，联合有关部门制定合同示范文本。根据前款规定制定的合同示范文本，应当主动公开，供社会公众免费阅览、下载、使用。(Article 13: Market supervision and administration departments at or above the provincial level may, in accordance with relevant laws and regulations, jointly formulate contract template texts for specific industries or fields with relevant departments. Contract template texts formulated under the preceding paragraph shall be proactively disclosed for free public access, download, and use.) \\
    & 第十四、十五、四十二条略。(Articles 14, 15, 42: omitted.) \\
    \midrule
    LLM Answer & 合同示范文本由国务院工商行政管理部门制定。(The contract template texts are formulated by the State Council's administrative department for industry and commerce.) \\
        \bottomrule
    \end{tabularx}
\end{table*}

\end{document}